%% file: main.tex
\documentclass[10pt,twocolumn,letterpaper]{article}

\usepackage[pagenumbers]{iccv} 
\usepackage{graphicx}
\usepackage{booktabs} 

\usepackage{comment}
\usepackage{pifont}
\definecolor{iccvblue}{rgb}{0.21,0.49,0.74}
\definecolor{salim}{rgb}{0.21,0.49,0.74}
\definecolor{clem}{rgb}{1,0.2,0.5}
\usepackage[pagebackref,breaklinks,colorlinks,allcolors=iccvblue]{hyperref}
\usepackage{graphicx}  
\usepackage{booktabs}  
\usepackage{xspace} 
\usepackage{multirow}  

\newcommand{\our}{\textbf{VIPER}\xspace}
\def\paperID{3882} 
\def\confName{ICCV}
\def\confYear{2025}

\title{VIPER: Visual Perception and Explainable Reasoning for Sequential Decision-Making}

\author{\textbf{Mohamed Salim Aissi\textsuperscript{1*}}
,
\textbf{Clémence Grislain\textsuperscript{1}}
,
\textbf{Mohamed Chetouani\textsuperscript{1}},\\
\textbf{Olivier Sigaud\textsuperscript{1}}
,
\textbf{Laure Soulier\textsuperscript{1}}
,
\textbf{Nicolas Thome\textsuperscript{1,2}}\\
\textsuperscript{1}Sorbonne Université, CNRS, ISIR, F-75005 Paris, France\\
\textsuperscript{2}Institut universitaire de France (IUF)\\
\textsuperscript{*} Corresponding author: mohamed-salim.aissi@Sorbonne-universite.fr
}

\begin{document}
\maketitle
\begin{abstract}

While Large Language Models (LLMs) excel at reasoning on text and Vision-Language Models (VLMs) are highly effective for visual perception, applying those models for  visual instruction-based planning remains a widely open problem. 
In this paper, we introduce \our, a novel framework for multimodal instruction-based planning that integrates VLM-based perception with LLM-based reasoning. Our approach uses a modular pipeline where a frozen VLM generates textual descriptions of image observations, which are then processed by an LLM policy to predict actions based on the task goal. We fine-tune the reasoning module using behavioral cloning and reinforcement learning, improving our agent's decision-making capabilities.
Experiments on the ALFWorld benchmark show that \our significantly outperforms state-of-the-art visual instruction-based planners while narrowing the gap with purely text-based oracles.  By leveraging text as an intermediate representation, \our also enhances explainability, paving the way for a fine-grained analysis of perception and reasoning components.   

\end{abstract}

\input{intro_sans_grounding}

\section{Related Work}
\label{sec:related_work}
\subsection{LLM and VLM for Sequential Decision-Making}
When tackling long-horizon multimodal sequential decision-making tasks, embodied agents leverage key internal mechanisms such as perception and reasoning to achieve their objectives. Recently, LLMs and  VLMs have demonstrated emergent capabilities in understanding complex situations, establishing them as essential architectures for embodied agents \cite{ahn2022icanisay, shinn2023reflexion, yao2023react}. However, these models face challenges related to grounding and alignment with the physical world \cite{mahowald2023dissociating}.
A primary line of research focuses on leveraging LLMs for sequential decision-making through text, either by filtering irrelevant actions for a given task~\citep{yao2023react,shinn2023reflexion} or by learning from environmental feedback via reinforcement learning ~\citep{carta2023grounding,tan2024true}. These methods have demonstrated impressive performance and generalization capabilities. However, they are primarily designed for textual environments,  where all state information is transmitted by the environment through text, limiting their applicability to real-world embodied agents.
Recently, research has increasingly focused on leveraging  VLMs for sequential decision-making tasks in visual environments. For instance, EMMA \cite{yang2024embodied} trained a VLM-based agent for sequential decision-making in visual environments by distilling the actions of a (SOTA) textual agents, such as Reflexion~\cite{shinn2023reflexion}, within a parallel textual world. This approach represents an initial step toward utilizing multimodality for instruction generation. However, compared to our work, it remains dependent on a parallel textual environment for task learning, limiting its applicability to real-world scenarios.
Closer to our work, RL4VLM \cite{zhai2024fine} employs reinforcement learning to train VLM-based agents within visual environments, enabling them to learn to achieve tasks across diverse environments. However, its performance remains inferior to methods that rely solely on textual observation for reasoning and planning. To address these limitations, we propose a framework that leverages the reasoning capabilities of LLMs in the textual modality and applies them to real-world scenarios through the vision modality. Our approach \our creates a decomposition between perception and reasoning via an intermediate text modality: a VLM component processes visual input to generate textual descriptions of all elements present in a scene, and a fine-tuned LLM utilizes this description to generate actions that achieve the specified goal.

\subsection{Text as Intermediate Modality}
Recently, VLMs have demonstrated impressive performance on mainstream computer vision tasks such as object detection, visual question answering (VQA), and image captioning \cite{ghosh2024exploring, zhang2024vision}. These tasks can be addressed using a single architecture and prompting, making VLMs a promising candidate for robotic applications. However, recent studies have shown that these architectures still underperform compared to their LLM counterparts on purely textual data. \citep{wang2024pictureworththousandwords} demonstrated that VLMs often exhibit inferior reasoning capabilities compared to LLMs, highlighting the advantages of text-based reasoning. For example, 
\citep{ozdemir2024enhancingvisualquestionanswering} utilized a VLM for image captioning before performing a question-answering task to enhance image-related information and improve VQA performance, demonstrating the role of VLMs in extracting scene information. Similarly, 
\citep{guran2024taskorientedroboticmanipulationvision} employed a VLM to extract bounding boxes and object information from a scene for robotic tasks, while \citep{dang2025planningvisionlanguagemodelsuse} proposed Image-to-PDDL, where a VLM extracts a representation in Planning Domain Description Language (PDDL) \citep{McDermott1998PDDLthePD} of the scene, which is then processed by an LLM to generate robotic actions, effectively leveraging the strengths of both architectures.

\noindent Our framework follows this line of research by combining the capabilities of VLMs and LLMs. \our is, to the best of our knowledge, the first approach for utilizing a VLM to extract scene information in text format and a fine-tuned LLM to reason over these descriptions in visual instruction-based planning.

\section{The \our Framework}
\begin{figure*}[!htbp]
    \centering
    \includegraphics[width=1.0\linewidth]{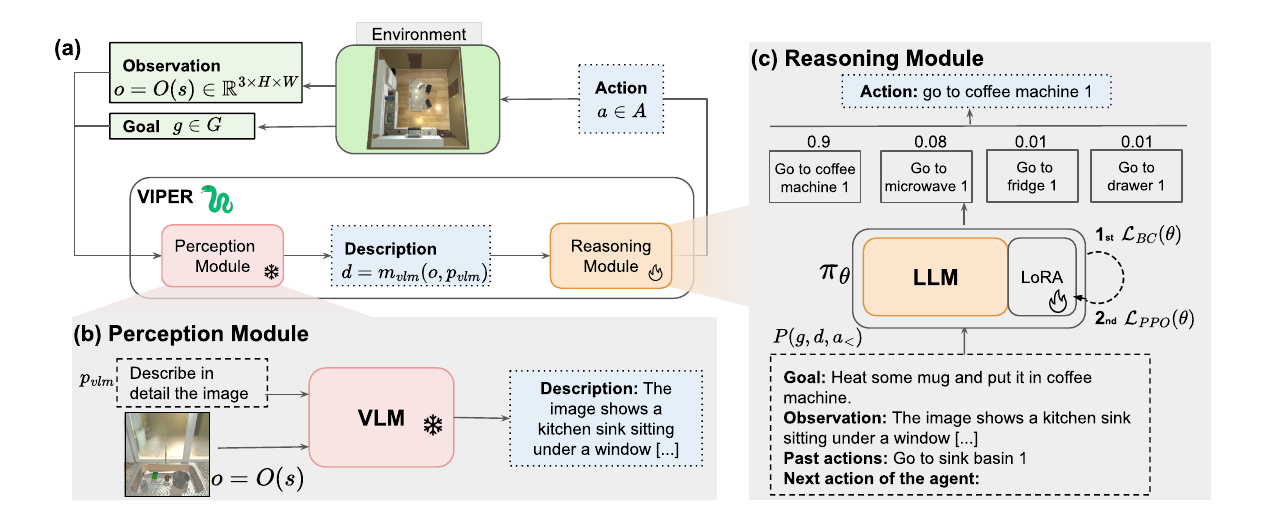}
    \vspace{-1cm}
    \caption{\textbf{The \our framework.} We employ a VLM and an LLM as respectively the perception and reasoning modules of the agent, to solve tasks in a household environment. \textbf{(a)} In this framework, the \textbf{perception module }generates a description of the observation, which is then fed to the reasoning module along with the task goal from the environment. The \textbf{reasoning module} then produces a textual action that the agent execute in the environment. \textbf{(b)} The \textbf{perception module } (VLM) generates a textual description of the image observation based on a fixed prompt.  \textbf{(c)} The \textbf{reasoning module} constructs a prompt by integrating the observation description, the goal, and past actions. This prompt is then fed into the LLM, which has been fine-tuned for decision-making. The LLM generates a probability distribution over possible actions and selects the most appropriate one based on the agent's current observation and its goal.}
    \label{fig:Main}
    \vspace{-0.5cm}
\end{figure*}



We introduce \our, a novel model for goal-oriented sequential decision-making in interactive multimodal environments. As shown in \figureautorefname~\ref{fig:Main}(a), our framework consists of two key components: (1) a \textbf{perception module}, which generates a textual description of the visual observation (Section~\ref{sec:preception_module} and \figureautorefname~\ref{fig:Main}(b)), and (2) a \textbf{reasoning module}, which processes this description alongside the goal and past actions to determine the next action (Section~\ref{sec:llm_decision} and \figureautorefname~\ref{fig:Main}(c)). We use a pre-trained VLM as our perception module and an LLM fine-tuned with a sequential training strategy combining BC and RL for our reasoning module (Section~\ref{sec:strategies} and \figureautorefname~\ref{fig:Main}(c)), allowing us to improve the grounding capabilities of the reasoning module on textual inputs while avoiding the need for textual supervision from the environment. While \our excels at solving sequential decision-making tasks, it also enables interpretable decision-making through the intermediate text modality between the perception and reasoning modules (Section~\ref{sec:interpretability}).

\subsection{Problem Statement}
\label{sec:problem}
We consider a goal-conditioned partially observable Markov decision process \( M = (S, A, T, R, G, O, \gamma) \), where \( S \) is the state space, $A$ is the action space, $G$ is the goal space, \( T: S \times A \mapsto S \) is the transition function, \( R: S \times G \mapsto \mathbb{R} \) is the goal-conditioned reward function and $\gamma$ the discount factor. More specifically, we consider a multimodal environment where, given a language vocabulary $\mathcal{V}$ and a maximum sequence length $N$, the goal $g\in G$ and the action $a\in A$ are in $\mathcal{V}^N$, while the observation of the state is an RGB image $ o = O(s) \in \mathbb{R}^{3 \times H \times W}$ where $H\times W$ are the image dimensions.

Our goal in this environment is to build an embodied multimodal agent, characterized by a policy $\pi$, that processes an image observation $o$ from the environment at each step and proposes an action $a \in A$ given the task goal $g$.

\subsection{State Description through VLMs}
\label{sec:preception_module}
To incorporate text as an intermediate modality for decision-making, we define a mapping function between image observation $o$ and its textual description $d=(w_1,w_2,...,w_N)\in D \subset \mathcal{V}^N$. Specifically, we leverage the strong captioning capabilities of VLMs to generate these textual descriptions and define a VLM mapping function conditioned on a prompt $p_{vlm}$:~$m_{vlm}(.|p_{vlm}):\mathbb{R}^{3 \times H \times W} \to D$. As detailed in Figure~\ref{fig:Main}(b), we use a fixed prompt, $p_{vlm}=$\textit{"Describe in detail the image"}, which remains agnostic to both the goal and the current step in the trajectory. The VLM is used exclusively for perception, while the decision-making process is entirely handled by the LLM, as detailed in Section~\ref{sec:llm_decision}. Further details on VLM prompt engineering can be found in Sup.~\ref{app:prompt_eng_vlm}.

\subsection{LLMs as Decision-Making Policies}
\label{sec:llm_decision}

We leverage the reasoning capabilities of LLMs to solve sequential decision-making tasks~\cite{carta2023grounding, shinn2023reflexion, tan2024true}. We follow the framework introduced in recent literature~\cite{carta2023grounding, tan2024true} to derive a policy $\pi$ (a probability distribution over possible actions) from an LLM, as shown in Figure~\ref{fig:Main}(c). More precisely, given a prompt \(p\), the probability of an action sequence \(a_i \in A\), denoted as \(\pi(a_i|p) = \mathbb{P}_{LLM}(a_i|p)\), is obtained from the decoding probabilities of the tokens \(\{w_j\}\) composing \(a_i\). Specifically, this probability is computed as:  

\begin{equation}
\label{eq:proba_action}
    \pi(a_i|p) = \frac{1}{|a_i|}\prod\limits_{j=0}^{|a_i|} \mathbb{P}_{LLM}(w_j | p, w_{<j}).
\end{equation}  

The recent state-of-the-art RL4VLM agent~\cite{zhai2024fine} uses free-form generation of actions, including possible actions in the prompt. 
~In contrast, the policy $\pi(a_i|p)$ in \cref{eq:proba_action} is directly constrained to the support of possible actions, improving performances on the downstream task (see \cref{sec:results}). 

In addition, while existing LLM agents methods~\cite{shinn2023reflexion,yao2023react,deps} receive prompts with textual observations directly from the environment, our method constructs the prompt using a textual description of the image observation generated by a VLM \(d = m_{vlm}(o|p_{vlm})\), along with the goal and the past actions.





\subsection{Training Strategies}
\label{sec:strategies}


Recent works \cite{carta2023grounding, aissi-etal-2025-reinforcement} show that while LLMs possess common-sense knowledge, they lack grounding and are misaligned with physical environments \cite{mahowald2023dissociating}, leading to poor zero-shot performance in complex interactive decision-making tasks. To address this, following \cite{carta2023grounding, tan2024true}, we fine-tune the reasoning module (LLM) using a sequential strategy. First, we align it with a good reference policy using BC on a dataset of expert demonstrations, which is assumed to be available in the environment. Then, we refine its performance through RL via exploration and interaction with the environment. This BC+RL approach is well-established in the control community \cite{lu2023imitation} and has demonstrated its relevance in our ablation study (see Section~\ref{sec:ablations}). While \our's reasoning module operates on textual inputs, the textual description of the image observation is provided by a frozen VLM. Therefore, unlike EMMA~\cite{yang2024embodied}, which trains its model by distilling a text expert policy and thus requires textual observations from the environment, our training relies solely on the goal and the image observations $o$. 

\par\textbf{BC fine-tuning:} We pre-train the LLM policy $\pi$ parametrized by a parameter $\theta$, using a dataset of expert demonstrations $\mathcal{T}:=\{({t}_i,g_i)\}^{\left| G \right|}_{i=1}$ of observation-action trajectories ${t_i}=\{o_0,a_0,o_1,a_1,...\}$ through BC. 
In our case, this expert dataset is created by selecting the successful trajectories of a rule-based expert (Sup.~\ref{app:ENV_SPEC}). The BC objective is to maximize the log-likelihood of expert actions with respect to the action distribution predicted by the LLM policy $\pi$ along the expert trajectory. Therefore, the BC loss is the negative log-likelihood, defined as:
\begin{equation}
    \mathcal{L}_{BC}(\theta)=-\sum_{{t_i}\in \mathcal{T}}\sum_{(o_j,a_j)\in {t_i}}log \left(\pi_\theta(a_j|P(g_i,d_j,a_{<j}))\right)
\end{equation}
where $d_j$ is the VLM description of the observation $o_j$, $d_j=m_{vlm}(o_j)$ and $P$ is the prompting function defined in Sup.~\ref{app:prompt_eng}.

\textbf{RL fine-tuning.} We further enhance the policy through online interaction with the environment and fine-tune \( \pi_\theta \) using Proximal Policy Optimization (PPO) \cite{schulman2017proximal}. We learn a value function \( \hat{V}: D \times G \mapsto \mathbb{R} \) that approximates the true value \( V(s, g) = \mathbb{E}_{a \sim \pi_\theta(a|s)}[R(s, g) + \gamma \times V(T(s, a), g)] \) for PPO. The policy \( \pi_\theta \) is then optimized to minimize \( \mathcal{L}_{PPO}(\theta) \), as described in~\citep{schulman2017proximal}.

\subsection{Explaining Agent Decision}
\label{sec:interpretability}

During inference, \our first executes the VLM to generate a textual description of the scene, which is then passed to the LLM to infer the action. This two-step prediction process enhances the explainability of \our's behavior, as the model's perception can be interpreted through the generated text. 
We can thus analyze the relevance of the text description in regard to the goal.

Furthermore, we perform weakly-supervised instance segmentation of important concepts in the image description used by the reasoning module by computing the integrated gradient~\cite{sundararajan2017axiomaticattributiondeepnetworks} of the chosen action $a^*$ with respect to each text token in the image description, and selecting the most significant ones. Following the same procedure, we compute the integrated gradient of the selected tokens, with respect to the image (see Sup.~\ref{app:interp_full} for more details). This provides a rich tool for explainability, linking the performance of the perception and reasoning modules, and enabling to finely analyze failure cases (see \cref{sec:expanalysis}).

\section{Experimental Results}
\label{sec:results}

\begin{table*}[t]
    \centering
        \begin{minipage}{0.95\textwidth} 
        \centering
        \resizebox{\textwidth}{!}{\begin{tabular}{lccccccc|c}
\midrule
\textbf{} & 
Environnement & 
Pick & Look & Clean & Heat & Cool & Pick2 & Avg  
\\ \midrule
AutoGen*~\citep{autogen}&  Text & 
0.92 (-) & 0.83 (-) & 0.74 (-) & 0.78  (-)& 0.86 (-) & 0.41 (-) & 0.77 (-) \\ 
ReAct*~\citep{yao2023react}&  Text & 
0.71 (18.1) & 0.28 (23.7) & 0.65 (18.8) & 0.62 (18.2) & 0.44 (23.2) & 0.35 (25.5) & 0.54 (20.6)\\ 
DEPS*~\citep{deps}&  Text & 
0.93 (-) & 1.00 (-) & 0.50 (-) & 0.80 (-) & 1.00 (-) & 0.00 (-) & 0.76 (-)\\ 
Reflexion*~\citep{shinn2023reflexion}&  Text & 
0.96 (17.4) & 0.94 (16.9) & 1.00 (17.0) & 0.81 (19.4) & 0.83 (21.6) & 0.88 (21.6) & 0.91 (18.7) \\ 
  \midrule
  EMMA*~\citep{yang2024embodied} & Text + Vision
 & 
0.71 (19.3) & 0.88 (19.6) & 0.94 (17.5) & 0.85 (19.6) & 0.83 (19.9) & 0.67  (22.4) & 0.82 (19.5) \\
 \midrule
 Florence-2~\cite{xiao2023florence}&  Vision
  &   
0.00 (30.0) & 0.06 (28.5) & 0.0 (30.0) & 0.0 (30.0) & 0.0 (30.0) & 0.0 (30.0) & 0.01 (29.7)  \\
 Idefics-2~\cite{laurençon2024matters}&  Vision
  &   
0.04 (29.2) & 0.06 (28.2) & 0.0 (30.0) & 0.0 (30.0) & 0.0 (30.0) & 0.0 (30.0) & 0.02 (29.5)  \\

 MiniGPT-4* \cite{zhu2023minigpt4enhancingvisionlanguageunderstanding} &  Vision
  & 
0.04 (29.0)  & 0.17 (17.7) & 0.0 (30.0) & 0.19 (26.3) & 0.17 (26.7) & 0.06(28.9) & 0.16 (26.9) \\ 
InstructBLIP* \cite{dai2023instructblipgeneralpurposevisionlanguagemodels} &   Vision
 & 
0.50 (21.5) & 0.17 (26.8) & 0.26 (25.0) & 0.23 (27.2) & 0.06 (28.9) & 0 (30.0) & 0.22 (26.2) \\  
RL4VLM*~\citep{zhai2024fine} &   Vision
 & 
0.47 (-) & 0.14 (-) & 0.10 (-) & 0.14 (-) & 0.18 (-) & 0.18 (-) & 0.21 (-) \\ 
\our&  Vision
 & 
0.80 (13.1) & 0.77 (16.7) & 0.77 (19.5) & 0.92 (14.3) & 0.71 (20.4) & 0.53 (24.0) & 0.75 (18.0) \\
 \midrule
\end{tabular}
}
        \caption{\label{tab:SOTA}: \textbf{Comparison with the state-of-the-art.} Mean success rate ($\uparrow$) and episode length ($\downarrow$) into brackets of various methods (*-reported in previous work). 
 and  respectively represent access to textual and visual observation from the environment. Florence-2, Idefics-2, MiniGPT-4 and InstructBLIP are used in zero shot. \textbf{\our surpasses SOTA  methods solely accessing visual observations and narrows the gap with oracle methods relying on textual observations.\vspace{-0.5cm}}
 }
        
    \end{minipage}
        \hfill

\end{table*}

 We conduct a series of experiments to evaluate our framework in sequential decision-making tasks within the ALFWorld benchmark. Our study aims to address the following research questions: $\bullet$ How does \our perform compared to state-of-the-art models? $\bullet$ What are the benefits of our approach in terms of explainability and performance with the integration of an intermediate text modality? $\bullet$ What is the impact of different architectural configurations in the perception and reasoning modules, as well as various training regimes, on overall performance?
 
\subsection{Experimental Setup}
\label{sec:study}

\noindent \textbf{Environment.} 
Our experiments are conducted on the ALFWorld benchmark~\cite{shridhar2021alfworldaligningtextembodied}, a 3D environment that simulates six categories of household tasks with textual goals. ALFWorld has become a standard benchmark in the literature for evaluating models in long-horizon multimodal decision-making tasks, offering a realistic and interactive environment well suited for studying real-world embodied applications. 
This benchmark contains sets of training, in-distribution validation, and out-of-distribution (OOD) episodes that encapsulate specific layouts of the environment (room in the house, object and furniture placement,  etc.). We follow the standard protocol and train \our on the training set of episodes, perform cross-validation on the in-distribution validation set
, and report our results on the OOD test set. We evaluate the models using the mean success rate (higher is better, $\uparrow$) and the mean length of successful trajectories (lower is better, $\downarrow$). See Sup.~\ref{app:ENV_SPEC} for more details about the environment.

\medbreak

\noindent \textbf{Architecture and Training.}
\our uses Florence-2-220M~\cite{xiao2023florence} as VLM and Mistral-7B~\cite{jiang2023mistral7b} as LLM, with further discussions on model choices in Section~\ref{sec:ablations}. We fine-tune the LLM as described in Section~\ref{sec:strategies} using low-rank adaptation (LoRA) with parameters $r=32$ and $\alpha=16$ on the attention layers. For BC, we train the model until convergence and for PPO for approximately \(5 \times 10^{5}\) interaction steps with the environment. Both fine-tuning processes use a batch size of 2 and the Adam optimizer with a learning rate of \(10^{-4}\). For each task, BC fine-tuning takes $\approx 20h$  and PPO $\approx 200h$ on 4 H100 GPUs.

\medbreak

\noindent \textbf{Baselines.} We compare \our with various SOTA models from the literature, categorized into two groups. The first group comprises models trained with supervision from a parallel textual environment, including Reflexion \cite{shinn2023reflexion}, DEPS~\cite{deps}, ReAct~\cite{yao2023react}, AutoGen~\cite{autogen} which only receive textual observation of the environment and EMMA~\cite{yang2024embodied} which uses both visual and textual observation during training but only access visual observation during inference. Like \our, the second group consists of models that exclusively rely on visual information from the environment, including zero-shot models (MiniGPT-4 \cite{zhu2023minigpt4enhancingvisionlanguageunderstanding}, InstructBLIP \cite{dai2023instructblipgeneralpurposevisionlanguagemodels}, Florence 2~\cite{xiao2023florence}, Idefics 2~\cite{laurençon2024matters}) and a VLM fine-tuned through BC and RL, RL4VLM \cite{zhai2024fine}.
All methods of the first group accessing the textual observations benefit from richer scene information, including the number of objects, object IDs within the description, and the correspondence between object IDs and actions. In contrast, the models of the second group, including the zero-shot perception module in \our, cannot perfectly generate this level of detail, as this information is internal to the environment and not directly observable from images. Therefore, the models of the first group can be seen as \textit{oracle} baselines.

\subsection{Comparison with the state-of-the-art}

We compare \our with the two baseline families on ALFWorld (see \tableautorefname~\ref{tab:SOTA}), evaluating both the mean success rate and the mean trajectory length required to achieve the goals. As observed, \our outperforms zero-shot visual-only methods, achieving an average 50pt improvement in success rate and an 8-step reduction in trajectory length across multiple tasks. It also surpasses RL4VLM \cite{zhai2024fine}, the SOTA method most closely related to our work that applies both BC and RL fine-tuning. This lower performance of RL4VLM~\cite{zhai2024fine} may be explained by the use of CoT reasoning with free generation for action prediction, which prevents it from being constrained to the action space. This leads to irrelevant predictions, as explained in Section~\ref{sec:llm_decision}. Another factor could be the absence of past action history in the agent's prompt, as discussed in Sup.~\ref{app:prompt_eng}. Besides, \our outperforms ReAct \cite{yao2023react} and matches the performance of AutoGen \cite{autogen} and DEPS \cite{deps}, which rely on textual observations from the environment. Moreover, \our approaches the performance of oracle methods such as EMMA~\cite{yang2024embodied} and Reflexion~\cite{shinn2023reflexion}, achieving more than 70\% success rate in the first five tasks and more than 50\% in the Pick2 task. The performance gap between Pick2 and the other tasks results from the increased complexity of manipulating multiple objects. Text-based methods, such as EMMA and Reflexion, do not face this challenge, as they have access 
\begin{figure}[t]
    \centering
    \includegraphics[width=0.95\linewidth]{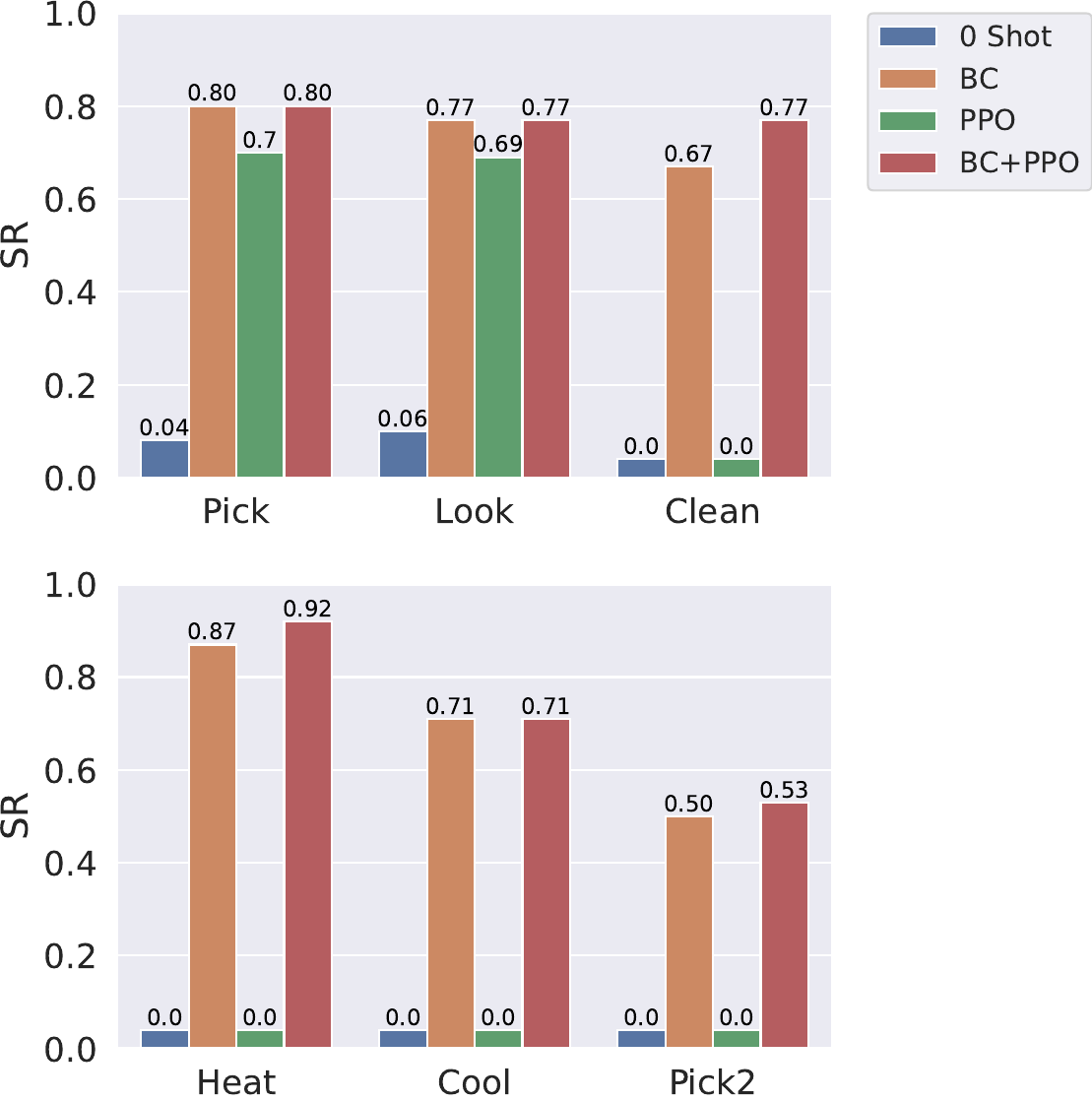}
    \vspace{-0.3cm}
    \caption{\textbf{Impact of BC and RL Training on Mean Success Rate.}
As observed, the combination \our BC+PPO yields the best overall performance in terms of success rate, highlighting the necessity of both BC and RL fine-tuning.}
    \label{fig:BCPPO}
    \vspace{-0.5cm}
\end{figure}
to internal environment information, including object identifiers and enumerations, which cannot be obtained solely through vision, as in \our (see Section~\ref{sec:study}).

Comparisons between textual observations and \our-generated descriptions can be found in Sup.~\ref{app:ENV_SPEC}. Furthermore, \our achieves the shortest mean episode length compared to both the oracle and vision-only baselines, demonstrating that its successful trajectories are more efficient, taking more goal-directed actions and avoiding unnecessary steps.
\begin{figure*}[t]
    \centering
    \includegraphics[width=0.90\linewidth]{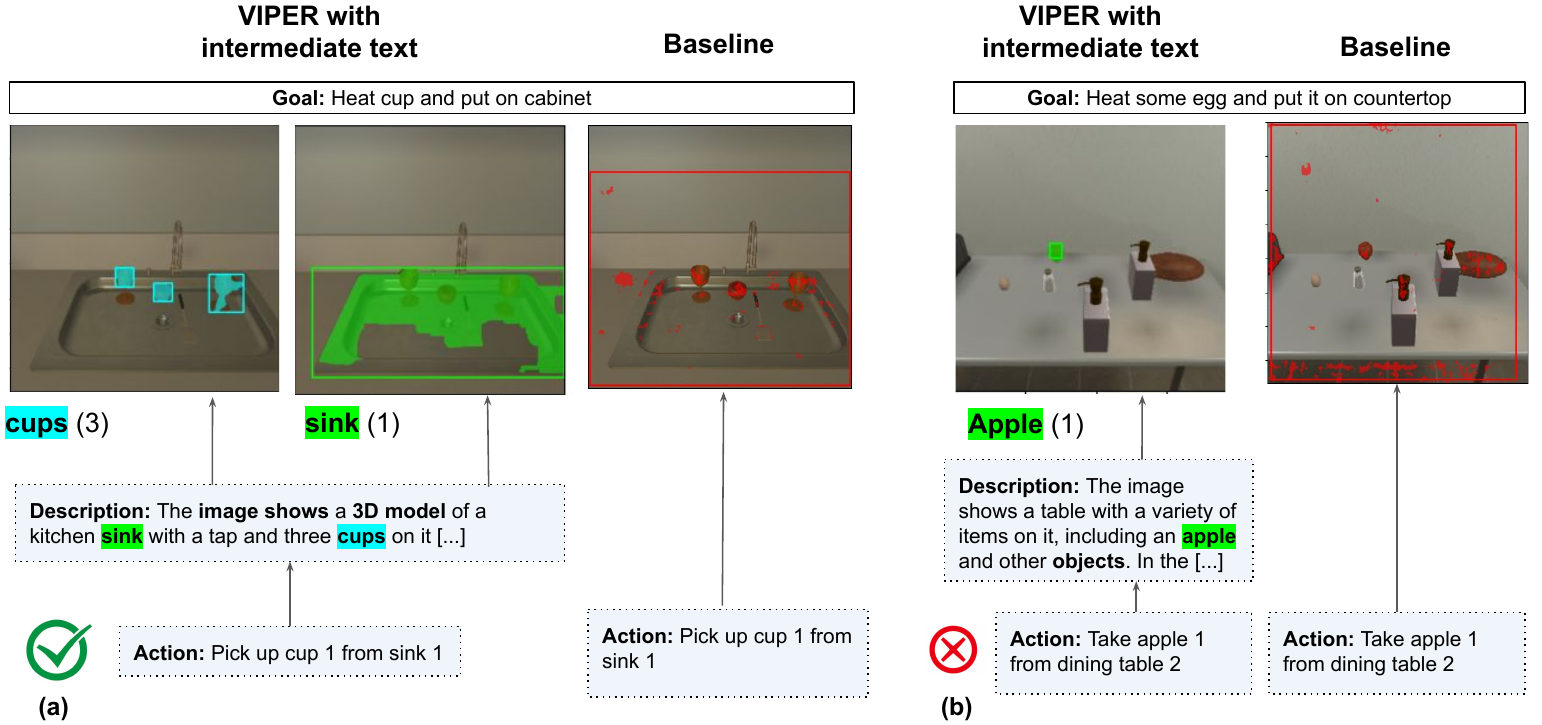}
    \caption{\textbf{Observation Analysis with Perception Modules.} We compare explainability in decision-making of Viper against a VLM agent baseline without intermediate text modality in examples of correct actions \includegraphics[height=1em]{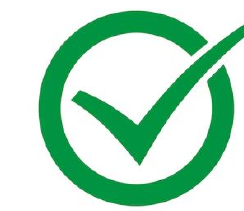} \textbf{(a)} and incorrect actions \includegraphics[height=1em]{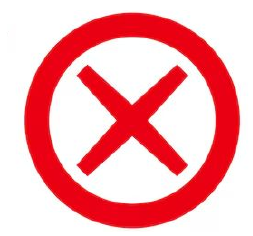} \textbf{(b)}. \our's intermediate text modality filters relevant elements for action generation and retrieves them from the image using integrated gradients, helping to analyze both correct (e.g., identifying cups and sinks in \textbf{(a)}) and incorrect predictions (e.g., missing the egg while focusing on the irrelevant apple in \textbf{(b)}).}
    \label{fig:interpretability}
    \vspace{-0.5cm}
\end{figure*}

\subsection{Ablation study}
\label{sec:ablations}

\noindent\textbf{Impact of BC and RL Training.}
Through an ablation study, we evaluate the impact of each training strategy defined in Section~\ref{sec:strategies} --~BC and RL~-- by comparing different variants of \our in \figureautorefname~\ref{fig:BCPPO}. Specifically, we report the performance of \our trained only with BC (\our BC) using rule-based expert demonstrations (see Sup.~\ref{app:ENV_SPEC}), only PPO (\our PPO), and the sequential approach where PPO follows BC (\our BC+PPO).
Although using SOTA VLMs and LLMs, the zero-shot model performs poorly across all tasks, highlighting the complexity of the tasks and the necessity of fine-tuning. The results further show that while applying RL fine-tuning alone (\our PPO) is sufficient to learn simpler tasks (Pick and Look), using BC beforehand is essential to learn more complex ones (Clean, Heat Cool and Pick2). Furthermore, while \our BC already outperforms vision-only SOTA models, adding RL further improves performance, producing a boost of 3pt to 9pt depending on the task. This performance gain is larger when considering weaker reasoning modules (see Sup.~\ref{app:training}).

\subsection{Impact of the Intermediate Perception Module}
\label{sec:expanalysis}
To assess the impact of the intermediate text modality in \our, we introduce a multimodal agent that follows the same prompt structure as our reasoning module but takes the image as input instead of the VLM-generated description. We refer to this baseline as the VLM agent.
\medbreak
\noindent\textbf{Improved Explainability.}
We now examine how the intermediate textual description generated by \our can be leveraged to analyze failures. \figureautorefname~\ref{fig:interpretability} presents two examples of action prediction: a successful action (a) and an unsuccessful one (b). The intermediate textual descriptions generated by the perception module of \our play a crucial role in understanding why the agent selects a correct or incorrect action.

In example (a), the integrated gradient of the description with respect to the action highlights elements like 'sink' and 'cups' as important, which helps to comfort the text description relevance with respect to the goal.  
Additionally, by computing the integrated gradient for these two concepts, we can extract their localizations in the image observation, further validating the correct behavior of the perception module.

In example (b), the agent selects an incorrect action w.r.t. the goal (it chooses an apple instead of an egg). Following the same procedure, we observe that the perception module generates the word apple in the description but does not include an egg. This indicates that the error originates from the perception module, which subsequently affects the reasoning process. Additional visualizations are provided in Sup.~\ref{app:interp_full}. In contrast, for the baseline VLM agent, computing the integrated gradient of the image with respect to the action results in noisy saliency maps and a bounding box covering the whole image, offering much weaker explainability.

\medbreak
\noindent\textbf{Impact of Intermediate Text on Performance.}
As we have just shown, introducing text as an intermediate modality between perception and reasoning improves Explainability. But it may also impact overall performance due to potential errors in perception.

We train two baseline VLM agents (first part of \tableautorefname~\ref{tab:vlm_llm_comparison}) based on the Idefics-2-8B~\cite{laurençon2024matters} and Florence-2-220M~\cite{xiao2023florence} models, using BC with the same dataset and number of iterations as \our. For a fair comparison and due to computational constraints, we omit the RL training step for the two VLM agents and \our.
As shown in \tableautorefname~\ref{tab:vlm_llm_comparison}, both \our and the VLM-based agents achieve comparable performance within the same parameter scale, with Florence/Mistral for \our and Idefics for the VLM agent. This shows that \our effectively combines the strengths of both worlds, achieving strong decision-making performance while enhancing explainability.

\begin{table}[t]
    \small  
    \centering
    \resizebox{\columnwidth}{!}{  
\begin{tabular}{c|ccccccc|c}
\midrule
Agent &  & Pick & Look & Clean & Heat & Cool & Pick2 & Avg \\ \midrule
\multirow{2}{*}{\begin{tabular}[c]{@{}c@{}}VLM \\ Agent\end{tabular}} & Idefics & \textbf{0.80} & 0.61 & \textbf{0.77} & \textbf{0.87} & \textbf{0.80} & 0.48 & \textbf{0.72} \\
 & Florence & 0.42 & 0.61 & 0.67 & 0.52 & 0.48 & 0.23 & 0.48 \\ \midrule
\multirow{3}{*}{VIPER} & Flo | Mistral & \textbf{0.80} & \textbf{0.77} & 0.67 & \textbf{0.87} & 0.71 & \textbf{0.53} & \textbf{0.72} \\
 & Flo | LLaMA & 0.62 & 0.61 & 0.67 & 0.74 & 0.40 & 0.35 & 0.57 \\
 & Ide | Mistral & \textbf{0.80} & 0.67 & 0.74 & 0.82 & 0.62 & \textbf{0.53} & 0.70 \\ \midrule
\end{tabular}
 
    } 
    \caption{\textbf{Impact of Intermediate Text modality and Model Analysis.} Upper part: VLM agent baselines, which operate without an intermediate text modality. Lower part: different variants of \our, each incorporating distinct perception and reasoning modules. All the agents are trained using\textbf{ solely BC fine-tuning.}}\vspace{-0.5cm}
    \label{tab:vlm_llm_comparison}
\end{table}

\subsection{Model Analysis}

\noindent\textbf{Impact of Perception and Reasoning Components.}
To evaluate the impact of each component in \our, we evaluate its performance with different perception (VLM) and reasoning (LLM) models of varying sizes: Idefics-2-8B and Florence-2-220M (respectively Ide and Flo in Table~\ref{tab:vlm_llm_comparison}) for perception, and Mistral-7B~\cite{jiang2023mistral7b} and LLaMA-1B~\cite{grattafiori2024llama3} for reasoning. The second part of Table~\ref{tab:vlm_llm_comparison} summarizes the results of the different combinations. We see that the reasoning module is the most critical component of \our, as the performance changes only of 2\% when swapping between the 8B Idefics and 220M Florence models for perception but drops from 72\% to 57\% in success rate on average over the six tasks when replacing Mistral-7B with LLaMA-1B for reasoning. These results confirm our initial assumption that while both perception and reasoning matter, reasoning is more critical, reinforcing our decision to fine-tune only the reasoning module.

\begin{figure}[t]
    \centering
    \includegraphics[width=0.92\linewidth]{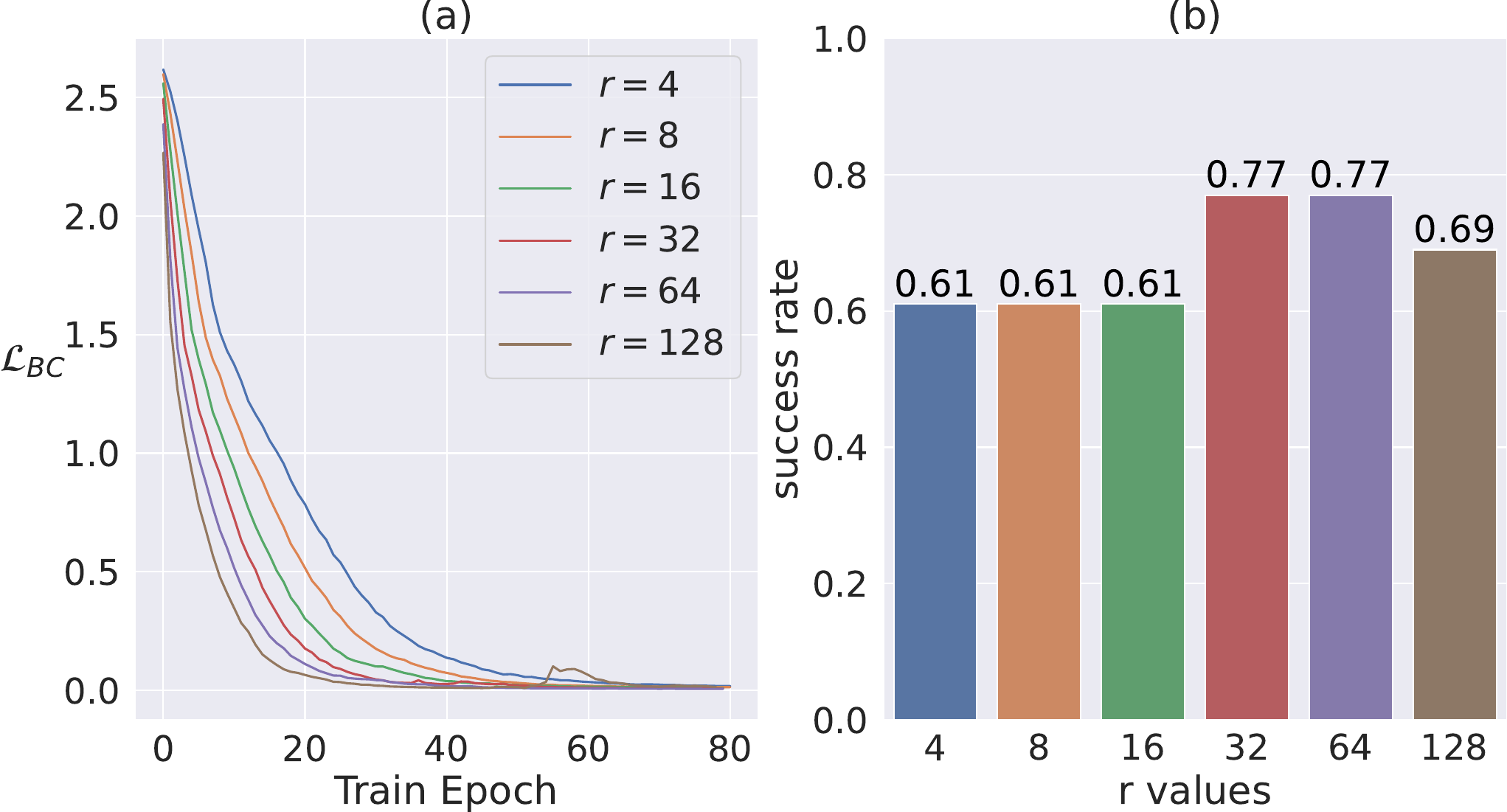}
    \caption{\textbf{LoRA Parameterization.} (a) Curves showing the performance of the BC framework trained with different LoRA parameters $r $.  
(b) Performance, in terms of success rate, for various $r$ values. $ r = 32 $ represents the optimal trade-off between performance and LoRA size.}
    \label{fig:lora}
    \vspace{-0.5cm}
\end{figure}
\medbreak
\noindent\textbf{LoRA Parameterization.} To fine-tune the pre-trained Mistral-7B reasoning module of \our, we use BC fine-tuning with LoRA, parameterized by the rank of the adapters $r \in [4, 8, 16, 32, 64, 128]$ and the update scaling factor $\alpha=\frac{r}{2}$ on the \textit{Look} task. Figure~\ref{fig:lora}~(a) shows the training curves. We observe that, as $r$ increases, the convergence speed improves, indicating that the model learns faster with a higher rank. Figure~\ref{fig:lora}~(b) shows the test set success rates. For $r<32$, the test set success rate is lower, suggesting that the rank is too low to capture the new knowledge required for generalizing sequential decision-making. For $r \geq 32$  the success rates stabilize, showing that the model has enough trainable parameters to effectively learn new knowledge. Therefore, we set $r=32$ and $\alpha=16$ in our final framework to maximize performance and minimize the number of trainable parameters.

\section{Conclusion}

In this paper, we introduce \our, a framework that integrates vision-language models (VLMs) and large language models (LLMs) for instruction-based planning. By leveraging VLMs for perception and LLMs for reasoning, \our enables an embodied agent to generate textual descriptions of visual observations and select actions based on instructions, improving grounding in real-world environments.  
Experiments on the ALFWorld benchmark show that \our enhances task performance compare to SOTA models accessing vision-only observations. Additionally, by generating human-readable descriptions, it facilitates human understanding and interaction with agents. These results suggest that \our can serve as a foundation for advanced decision-making frameworks.  

\noindent Future works include extending our approach 
~to low-level control in robotics, enabling
~fine-grained interactions with the environment and executing real-world tasks with greater autonomy and precision.
\section*{Acknowledgments}
Experiments presented in this paper were carried out using the HPC resources of IDRIS under the allocation 2025-[AD011015093R1] made by GENCI. This work was supported by the European Commission's Horizon Europe Framework Programme under grant No 101070381 (PILLAR-robots), by RODEO Project (ANR-24-CE23-5886), by PEPR Sharp (ANR-23-PEIA-0008, ANR,
FRANCE 2030) and by Cluster PostGenAI@Paris (ANR-23-IACL-0007, FRANCE 2030). We would also like to thank Clement Romac and Thomas Carta for the various exchanges held during the project.

{
    \small
    \bibliographystyle{ieeenat_fullname}
    \bibliography{main}
}

\input{sec/X_suppl}

\end{document}

%% file: intro_sans_grounding.tex
\section{Introduction}
\label{sec:intro}




The impressive capabilities of Large Language Models (LLMs) and Vision-Language Models (VLMs) are now well-demonstrated in various tasks, including respectively translation, summarization, and question answering \citep{singhal2023towards,yao2024tree,wei2023chainofthought} for Natural Language Processing (NLP) and object detection, image captioning, or visual question answering \citep{ghosh2024exploring,zhang2024vision} in computer vision. 

\begin{figure}[t]
    \centering
    \includegraphics[width=\linewidth]{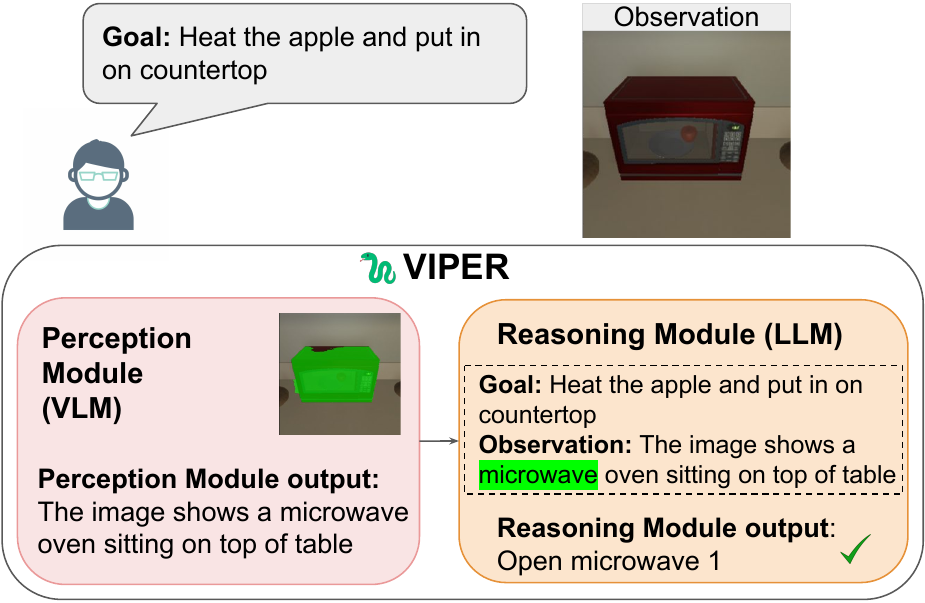}
    \caption{\textbf{VIPER framework for visual instruction-based planning}: from image observations and a user's goal, \our interacts with the environment to achieve it. \our is the first method to split the decision-making process using text as intermediate representation. The VLM-based perception module generates a textual scene description, and the LLM reasoner then selects the most appropriate action. By detecting the most important text tokens and image regions (green in Figure), \our has significant potential for explaining decision-making mechanisms.  }

    \vspace{-0.5cm}
    
    \label{fig:DNA}
\end{figure}

These last years, a new line of research leverages recent advancements in LLMs and VLMs for the development of embodied AI agents~\citep{zhao2024large,yao2023react}. One example of task, tackled in this paper, concerns visual-based instruction-based planning, where an agent is tasked with solving a problem formulated through a textual prompt (\eg, ``heat the apple and put it on the countertop") while using images as observations of the world — see \figureautorefname~\ref{fig:DNA}.
The direct application of LLMs and VLMs for solving instruction-following tasks remains far from trivial. In particular, the abstract knowledge embedded in LLMs/VLMs often lacks the experiential foundation needed to develop common sense and a grounded understanding of the world~\citep{huang2023grounded,ahn2022icanisay}. This grounding capacity requires additional skills, such as learning objects' physical properties and affordances, understanding possible interactions between them, and grasping the temporal causality of the world.




For text-based instruction-based planning, attempts have been made to enhance the relevance of LLMs actions, including prompting strategies such as In-Context Learning (ICL) and Chain-of-Thought (CoT) reasoning~\citep{shinn2023reflexion,yao2023retroformer}, as well as the integration of dedicated grounding modules~\citep{driess2023palm,jiang2022vima,li2023mastering}.
More recently, reinforcement learning (RL) has also emerged as a powerful method that enables models to acquire real-world knowledge through interaction with the environment~\citep{aissi-etal-2025-reinforcement,carta2023grounding,tan2024true}. When turning to visual-based instruction-based planning, recent works have tried to apply similar techniques to VLMs~\citep{zhai2024fine,yang2024embodied} by engaging in realistic multimodal environments and fine-tuning using RL. However, the results obtained in~\citep{zhai2024fine} remain significantly lower compared to those achieved with text-based models, while~\citep{yang2024embodied} relies on detailed text-based descriptions of each image, including object IDs, to guide the VLM during training—a cumbersome and unrealistic requirement in interactive environments.



This paper proposes a method for visual-based instruction-based planning that only requires images as world's observation, during both training and inference. 
We introduce the \our\footnote{code can be found \href{https://anonymous.4open.science/r/VIPER-Visual-Perception-and-Explainable-Reasoning-for-Sequential-Decision-Making-4773/README.md}{here} } framework, which integrate the \textbf{Vi}sual \textbf{P}erception capabilities of VLMs with the \textbf{E}xplainable text \textbf{R}easoning of LLMs. 


\our follows a modular approach, as illustrated in Figure~\ref{fig:DNA}: (1) a \textbf{perception module}, which generates textual goal-agnostic descriptions from image observations using a frozen VLM, and (2) a \textbf{reasoning module}, which predicts actions based on these descriptions, the task goal, and past interactions. The LLM acts as a policy model, deriving an action distribution from its token probabilities. This structured approach enables explainable, multimodal decision-making, bridging perception and reasoning through text for instruction-based planning.


\noindent In summary, our key contributions are as follows: 

\noindent $\bullet$ \our is the first method to leverage text as an intermediate representation between perception and reasoning in visual instruction-based planning. We integrate VLM-based perception with LLM-based reasoning, where the VLM describes the image observation to the LLM reasoner, which then generates the appropriate action based on a given textual goal.


\noindent $\bullet$ We fine-tune the LLM-based reasoning module using a combination of behavioral cloning (BC) with expert trajectories and RL from environment interactions. By leveraging a zero-shot VLM-based perception module to caption image observation, we enhance the decision-making capabilities of the reasoning module without the need for textual supervision from the environment. 

\noindent$\bullet$ Our intermediate text representation
~enables the development of rich explainability techniques.
~By detecting important tokens in the image description and the corresponding image regions, \our can perform weakly supervised instance segmentation, facilitate a fine-grained analysis of the decision-making process
, and provide tools for analyzing failure modes.



Extensive experiments conducted on the challenging ALFWorld environment~\citep{shridhar2021alfworldaligningtextembodied} reveal that \our significantly outperforms state-of-the-art visual instruction-based planning methods while narrowing the gap with respect to LLM-based oracles that use precise text descriptions as input. The relevance of our intermediate text representation for explainability is further validated by various quantitative results, while ablation studies validate the importance of our architectural and training choices. Source code will be released upon acceptance.


%% file: sec/X_suppl.tex
\clearpage
\setcounter{page}{1}
\setcounter{section}{0}
\maketitlesupplementary
\renewcommand{\thesection}{\Alph{section}}

\section{Environment Specifications}
\subsection{ALFWord Dataset}

In this section, we provide a detailed description of the data available in the ALFWorld environment. ALFWorld consists of six categories of household tasks, namely: "Pick and Place" (Pick), "Examine in Light" (Look), "Heat and Place" (Heat), "Cool and Place" (Cool), and "Pick2 and Place" (Pick2). The dataset is divided into four subsets: "Train", "Validation", and "Test (Out-of-Distribution)".  

Table~\ref{tab:dataset_stats} presents a summary of the number of instances for each task. Additionally, ALFWorld includes a rule-based expert for both the visual and textual environments, designed to solve the various tasks in the training set. In particular, the textual rule-based expert shows superior performance compared to its visual counterpart, as shown in Table~\ref{tab:dataset_stats}.
\begin{table}[!htbp]
    \centering
    \renewcommand{\arraystretch}{1.3}
    \setlength{\tabcolsep}{5pt}
\resizebox{\columnwidth}{!}{
\begin{tabular}{cccccc}
\hline
\multicolumn{1}{c|}{\multirow{3}{*}{Task}} & \multicolumn{3}{c|}{Train} & \multirow{3}{*}{\begin{tabular}[c]{@{}c@{}}Valid \end{tabular}} & \multirow{3}{*}{\begin{tabular}[c]{@{}c@{}}Test \\ OOD\end{tabular}} \\
\multicolumn{1}{c|}{} & \multirow{2}{*}{\begin{tabular}[c]{@{}c@{}}Total \\ data\end{tabular}} & \multirow{2}{*}{Labeled} & \multicolumn{1}{c|}{\multirow{2}{*}{Unlabeled}} &  &  \\
\multicolumn{1}{c|}{} &  &  & \multicolumn{1}{c|}{} &  &  \\ \hline
Pick and Place & 790 & 783 & 7 & 46 & 24 \\
Examine in Light & 308 & 303 & 5 & 19 & 18 \\
Clean and Place & 650 & 627 & 23 & 37 & 31 \\
Heat and Place & 459 & 438 & 21 & 25 & 23 \\
Cool and Place & 533 & 497 & 36 & 28 & 21 \\
Pick2 and Place & 813 & 734 & 79 & 45 & 17 \\ \hline
\end{tabular}}
    \caption{Number of episodes per task and the count of successful (labeled) and unsuccessful (unlabeled) trajectories generated by the rule-based expert in ALFWorld. Only the successful trajectories are used for BC fine-tuning.}
    \label{tab:dataset_stats}
\end{table}
\subsection{ALFWord Textual description}
\begin{figure*}
    \centering
    \includegraphics[width=\linewidth]{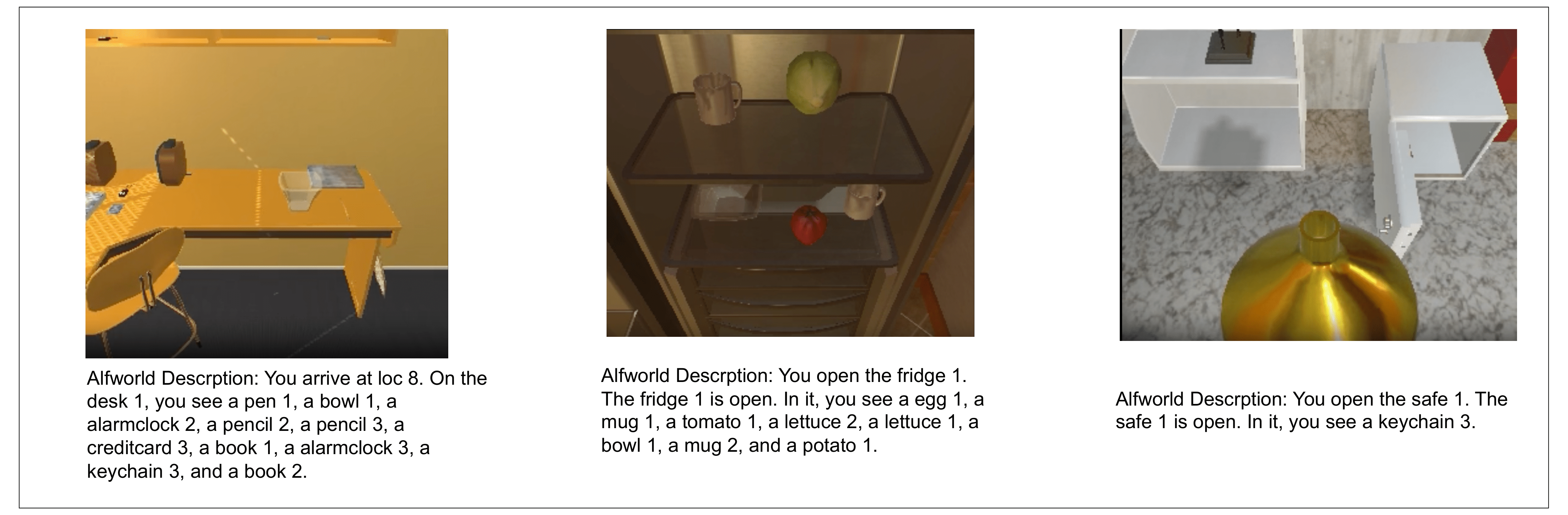}
    \caption{{\bf Visual and Textual Representations in ALFWorld.} An illustration of an image from the ALFWorld environment alongside its corresponding textual description, which provides detailed observations of the current state.}
    \label{fig:alfwordtext}
\end{figure*}

The ALFWorld environment includes a parallel textual environment that provides a textual description of observations and the current state of the environment. This textual environment offers internal information that is not available in the visual environment, such as object identifiers corresponding to actions and objects that are not visible in the image. This discrepancy explains the performance gap between text-based and vision-based agents, as observed in Table~\ref{tab:SOTA}. Figure~\ref{fig:alfwordtext} presents an example of a textual description generated by the environment.
\label{app:ENV_SPEC}

\section{Prompt Definition and Engineering}
\subsection{VLM Prompt Engineering}
\label{app:prompt_eng_vlm}

\begin{figure}[!htbp]
    \centering
    \includegraphics[width=\linewidth]{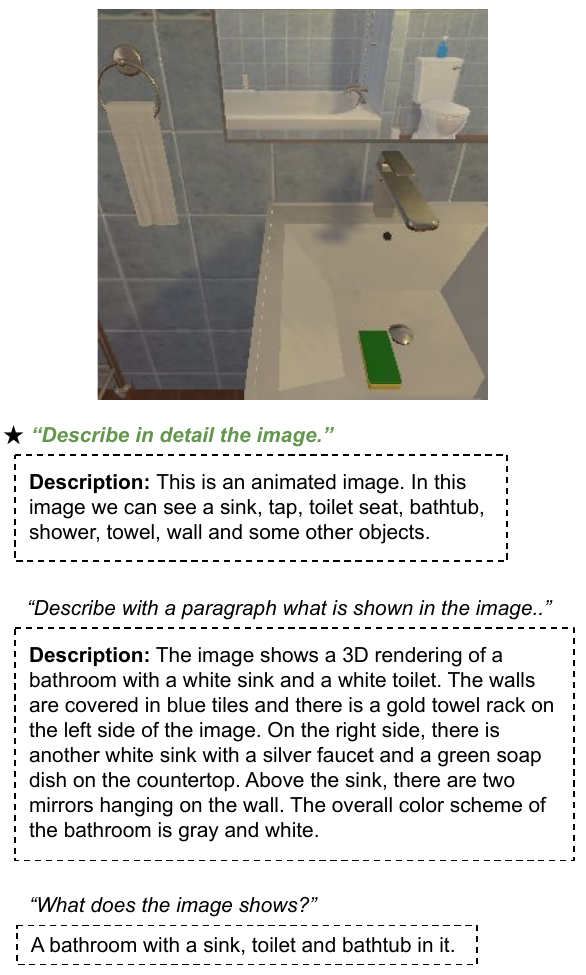}
    \caption{\textbf{VLM-Generated Descriptions with Different Prompts.} Descriptions generated by Florence-2-220M using various prompts. The prompt chosen for \our is highlighted in green, as it provides the most concise yet exhaustive description of the image observation.}
    \label{fig:prompt_vlm}
\end{figure}

In \our, the perception module (VLM) provides a goal-agnostic description of the current image observation from the environment. The reasoning module (LLM) then uses this description to determine the next action needed to achieve the goal. Therefore, descriptions should be concise, accurate, and free from hallucinations, focusing only on factual elements rather than subjective details like the room's style or atmosphere.

Based on a qualitative evaluation of different prompts using these criteria, we selected $p_{vlm}=$ \textit{"Describe in detail the image"} when using Florence as \our perception module and $p_{vlm}=$ \textit{"Describe in detail the image and all objects present in it."} when using Idefics in the experiments of Section~\ref{sec:ablations}. Figure~\ref{fig:prompt_vlm} illustrates the descriptions generated by Florence for the same observation with different prompts. Additionally, to reduce hallucinations, we generate textual descriptions using greedy decoding instead of sampling and apply a repetition penalty of value $1.08$ and a skip-ngrams penalty with parameter $n=3$.

\subsection{LLM Prompt definition}
\label{app:prompt_eng}

\begin{figure}[!htbp]
    \centering
    \includegraphics[width=\linewidth]{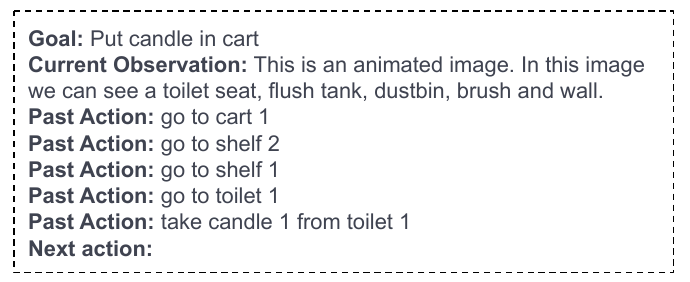}
    \caption{\textbf{Example of the Prompt Structure Used by the Reasoning Module (LLM)}}
    \label{fig:prompt_llm}
\end{figure}

The reasoning module is responsible for selecting the appropriate action based on the goal, the VLM-generated description of the image observation, and past actions. We incorporate past actions based on the findings of the EMMA ablation study \cite{yang2024embodied}, which demonstrates that access to past actions is critical to achieving the goal. Specifically, the ALFWorld tasks require exploring the environment to find relevant objects and furniture for the goal. Since the agent is subject to partial observability, recalling past actions is essential for effective decision-making. To balance memory with prompt size, we chose to include a maximum of 5 past actions in our prompt. Building on the work in \cite{aissi-etal-2025-reinforcement}, we use tags for key elements in the prompt—‘Goal,’ ‘Current Observation,’ ‘Past Action,’ and ‘Next Action’—with the LLM filling in the final line to propose the next action. An example of this prompt structure is shown in Figure~\ref{fig:prompt_llm}.

\section{Training Details and  Curves}

\begin{figure}[!htbp]
    \centering
    \includegraphics[width=0.9\linewidth]{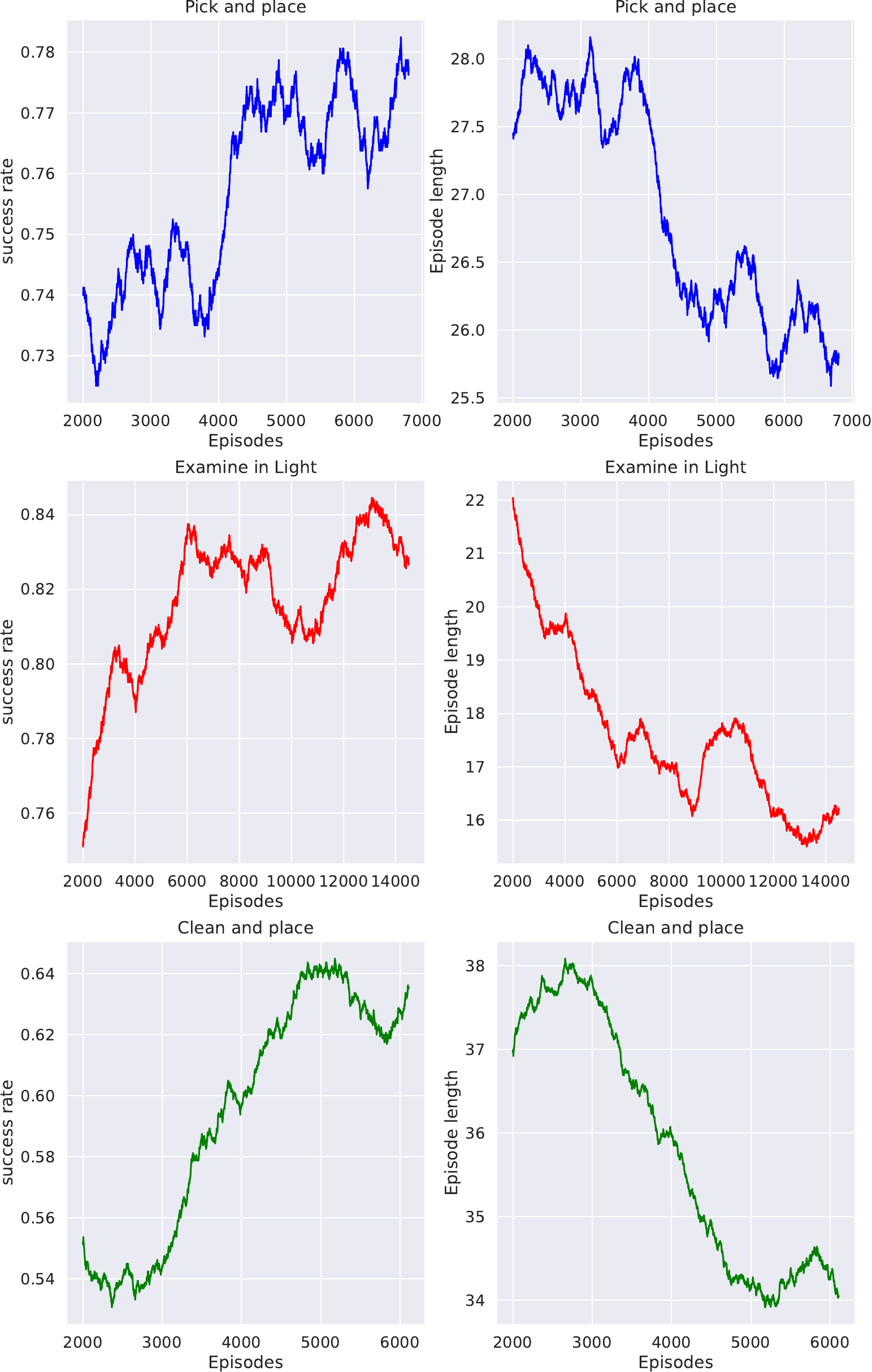}
    
    \caption{RL training curves of success rate and episode length for the tasks: Pick and Place, Examine in Light, and Clean and Place.}
    \label{fig:rl1}
\end{figure}

\begin{figure}[!htbp]
    \centering
    \includegraphics[width=\linewidth]{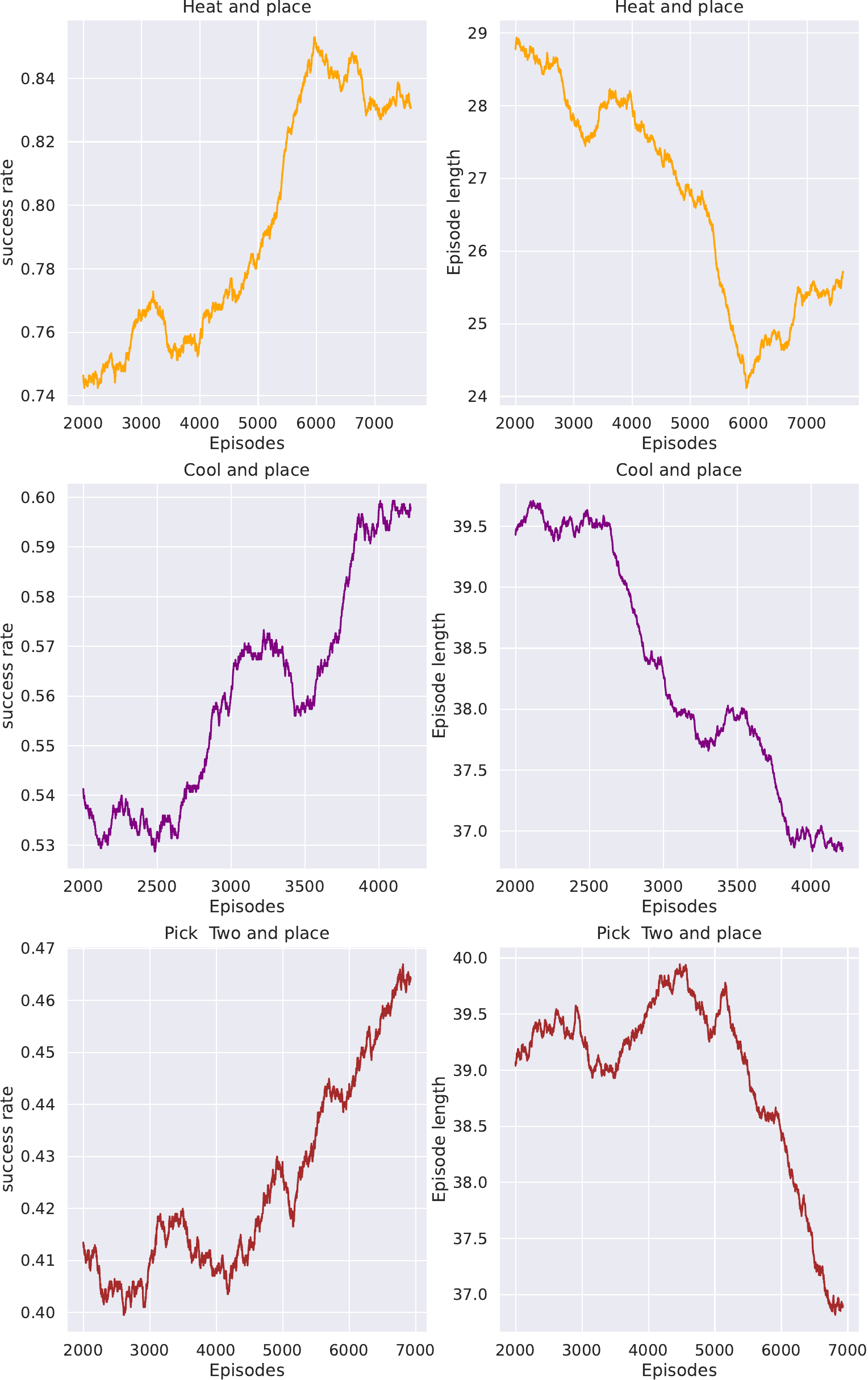}
    
    \caption{RL training curves of success rate and episode length: Heat and Place, Cool and place, and Pick two and Place.}
    \label{fig:rl2}
\end{figure}

\noindent\textbf{PPO training} Figure \ref{fig:rl1} and Figure~\ref{fig:rl2} illustrate the evolution of the success rate and episode length during PPO training. We see that online interaction with the environment improves the agent's performance by enabling exploration of new states and interactions. Furthermore, the duration of episodes decreases progressively with interactions, indicating that the LLM learns more efficient trajectories during RL training.

\begin{figure}[!ht]
    \centering
    \includegraphics[width=\linewidth]{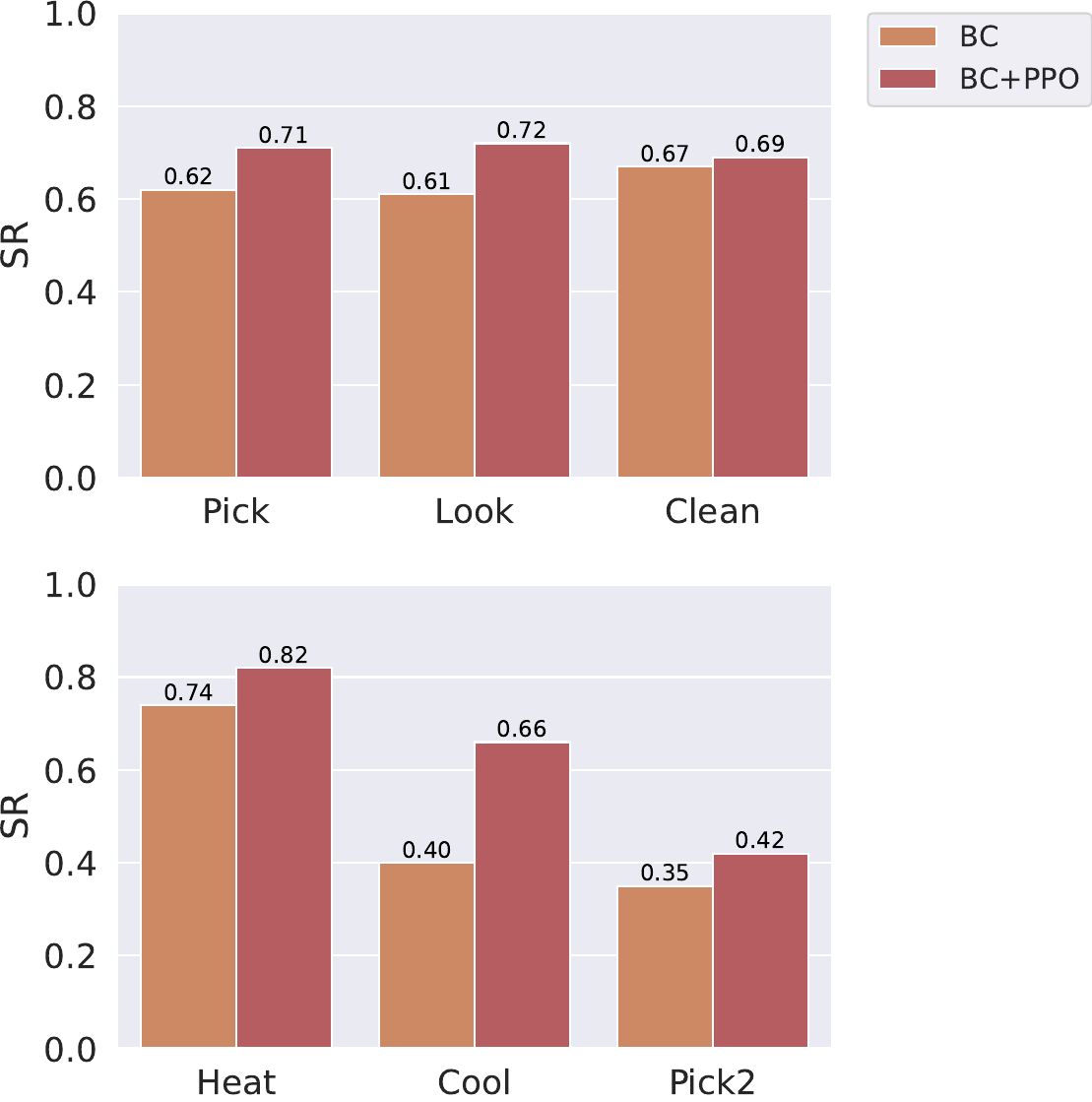}
    
    \caption{Performance of \our with Florence VLM and Llama LLM trained with BC fine-tuning alone and with BC followed by RL fine-tuning, over the six tasks, in terms of mean success rate.}
    \label{fig:llama_sup}
\end{figure}

For the best combination of \our (Florence and Mistral), PPO training improves success rates by 3 to 9 percentage points across tasks. However, as shown in Figure~\ref{fig:llama_sup}, when starting from a weaker initial model trained solely with BC (Florence with LLaMA), the improvement is more significant (improvement between 2pt and 26pt). This suggests that as the performance of the BC-trained model increases, further gains from PPO fine-tuning become smaller, indicating a saturation effect where further fine-tuning yields smaller improvements..

\begin{figure}[!htbp]
    \centering
    \includegraphics[width=\linewidth]{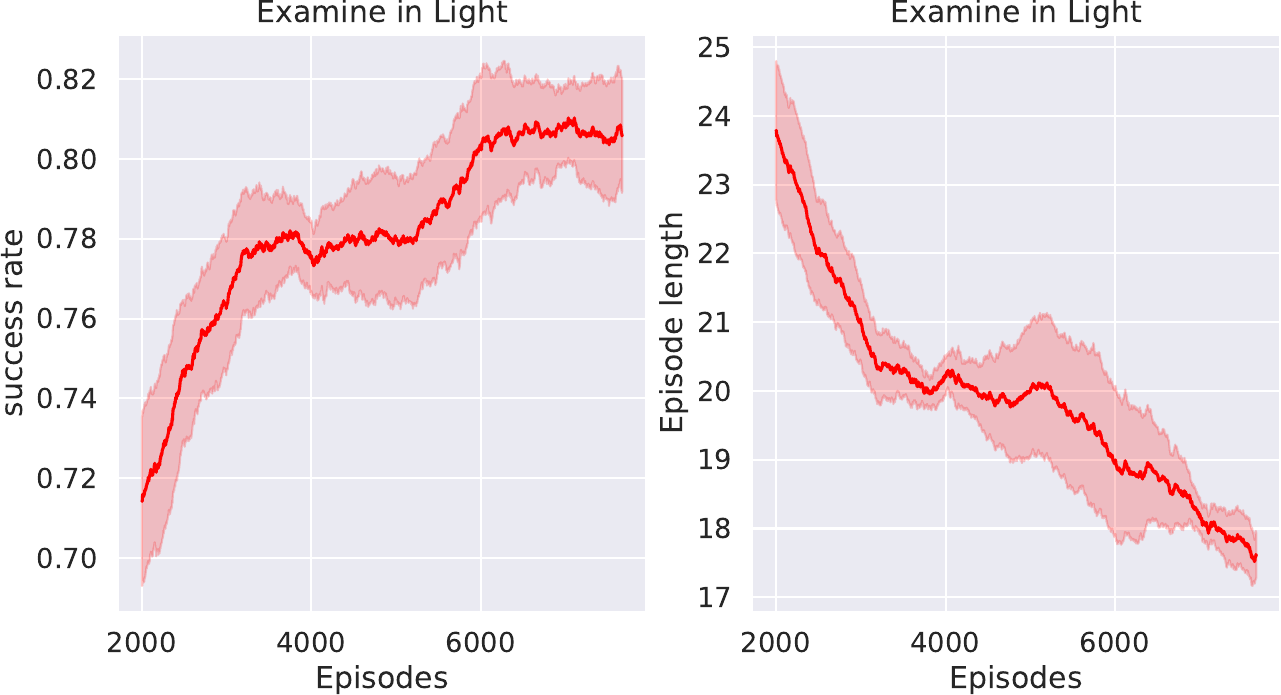}
    
    \caption{RL training curves of success rate and episode length for the task Examine in Light with three random seeds.}
    \label{fig:seed}
\end{figure}

\medbreak
\noindent\textbf{Results Variability} To assess the robustness of our models, we train with three different seeds on the Examine Light task. The results, presented in Figure~\ref{fig:seed}, show that the models follow a similar trend during training, with a standard error of 3\% in success rate and a standard error of 1 step in episode length.
\label{app:training}

\newpage
\section{Explainability: full interaction scenarios}
\label{app:interp_full}

\begin{figure*}[!htbp]
    \centering
    \includegraphics[width=\linewidth]{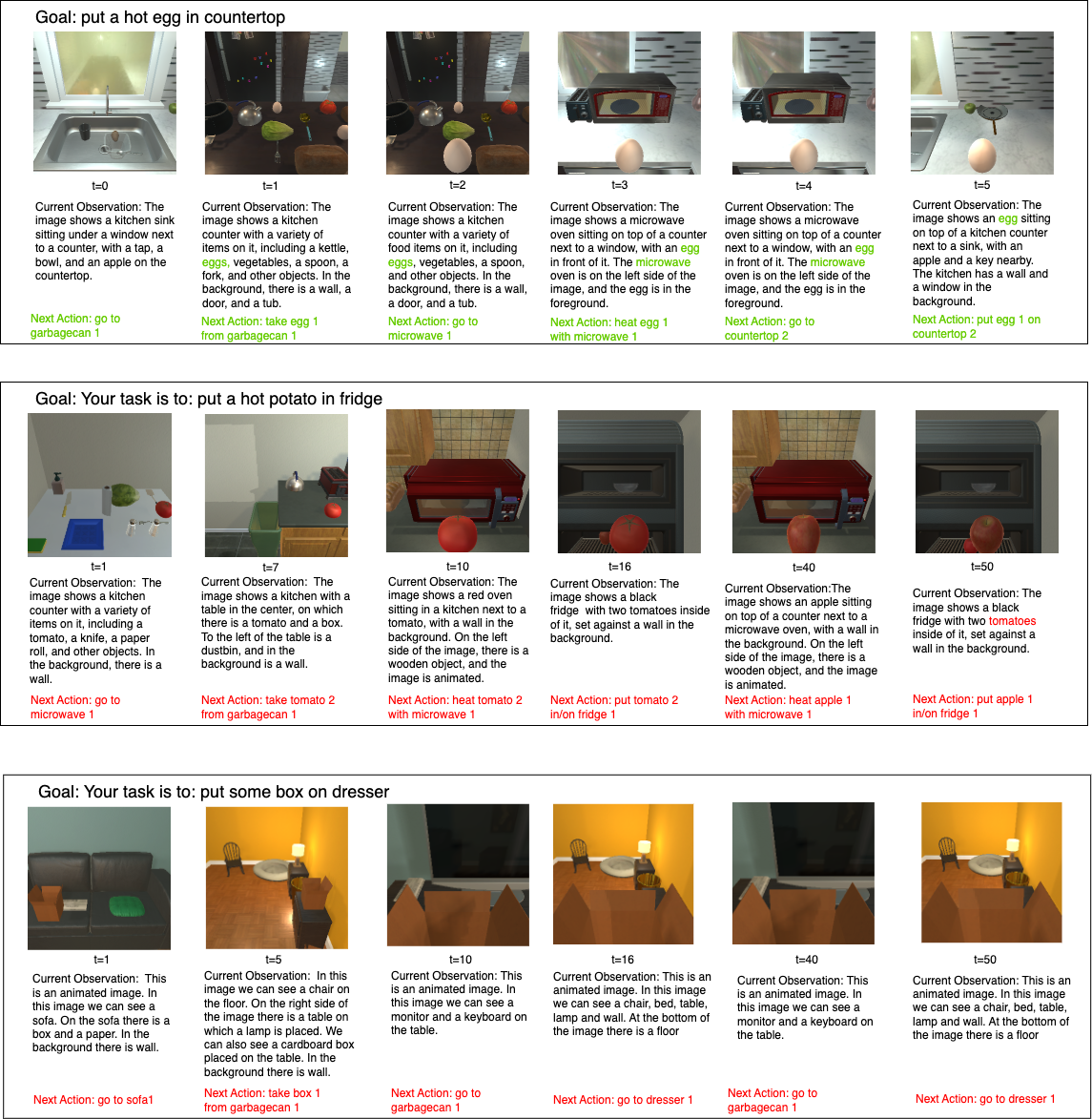}
    \caption{\textbf{Failure Analysis with Perception Modules}. The first row (a) shows a successful trajectory, while the next rows (b) and (c) illustrate failure cases. In (a), the perception module provides accurate descriptions for each frame, correctly identifying key elements needed to complete the task (e.g., eggs and microwave, highlighted in green). In (b), however, the agent fails to detect the potato in the scene, leading to incoherent actions that do not align with the goal. Instead, it mistakenly interacts with unrelated objects, such as tomatoes or apples. In (c), although the trajectory is unsuccessful, the text description confirms that the perception module correctly identified and mentioned the box. This suggests that the failure is due to a reasoning error from the LLM rather than a perception issue.}
    \label{fig:failure_full}
\end{figure*}

\begin{figure*}[!htbp]
    \centering
    \includegraphics[width=\linewidth]{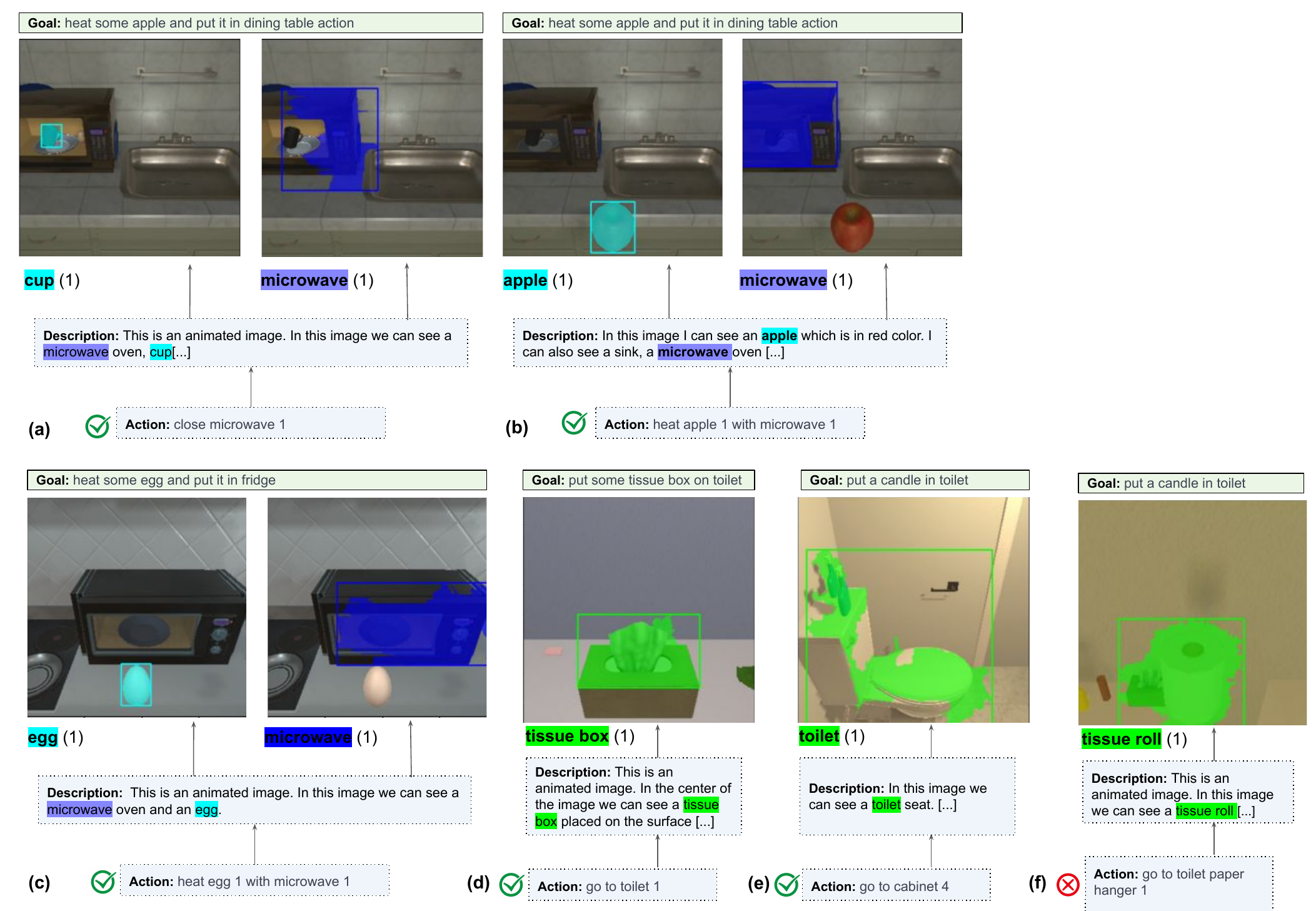}
    \caption{\textbf{Visualization of Key Words and Corresponding Visual Regions in Decision-Making}
Examples from the \textit{pick and place} (blue) and \textit{heat and place} (green) tasks illustrating $\our$'s interaction steps in the ALFWorld environment. Using the integrated gradients method (Section~\ref{sec:interpretability}), we highlight the most relevant words in the VLM-generated description for the selected actions and propagate this information to the images, marking the corresponding visual regions in the observations. In all these examples, the VLM accurately identifies relevant objects for the task, making the reasoning module responsible for the relevance of the actions, labeled as \includegraphics[height=0.8em]{sec/figures/good_icon.pdf} when correct and \includegraphics[height=0.8em]{sec/figures/bad_icon.pdf} when incorrect.}
    \label{fig:obs_sup}
\end{figure*}

Figure~\ref{fig:failure_full} illustrates how the intermediate text modality helps explain why the agent succeeds or fails in selecting the correct action for the goal. 

In the first example, the intermediate text description shows that the VLM accurately identifies and mentions the objects in the scene, enabling the LLM to successfully complete the task. In contrast, in the second example, the VLM fails to identify the potato on the table. As a result, the LLM selects an irrelevant action, since the goal 'put a hot potato in the fridge' requires finding the potato in the scene, but the agent mistakenly chooses to interact with the tomato instead. In such examples, the intermediate text description helps identify a perception mistake from the VLM and a reasoning mistake from the LLM which should engage in exploration to find the potato rather than interact with the tomato or other irrelevant objects for the goal. 
In the third example, the text description confirms that the perception module correctly identifies all objects in the scene, including the box, which is relevant for the goal. However, the selected action is incorrect, choosing to "go to the sofa" despite knowing that a box is present and that the goal is to "put some box on the dresser", ultimately leading to task failure. In this case, the selection of an incorrect action is purely due to a reasoning error from the LLM, which chooses to "go to the sofa" despite knowing that a box is present and that the goal is to "put some box on the dresser."

Figure~\ref{fig:obs_sup} demonstrates how our explainability framework (Section~\ref{sec:interpretability}) allows us to additionally identify the key concepts in the text description that influenced action selection in the reasoning module. Specifically, in these examples, the VLM accurately identifies the objects in the image. The error is therefore attributed to a reasoning mistake from the LLM, rather than a perception error from the VLM, as shown in Figure~\ref{fig:obs_sup}~\textbf{(f)}. In \textbf{(f)}, the reasoning module appears to overlook the goal and to incorrectly choose the action \textit{go to toilet paper hanger 1}—~as it holds a toilet paper roll~— rather than selecting actions aligned with the goal \textit{put candle in toilet}. In contrast, examples \textbf{(c)} and \textbf{(d)} show the LLM taking actions directly relevant to the goal, as in example \textbf{(a)}, as the action \textit{heat [...]} cannot be performed while the microwave door is open. In example \textbf{(e)}, the LLM engages in exploration to find relevant objects in the room, making the selected action aligned with the goal.

\begin{figure*}[!htbp]
    \centering
    \includegraphics[width=\linewidth]{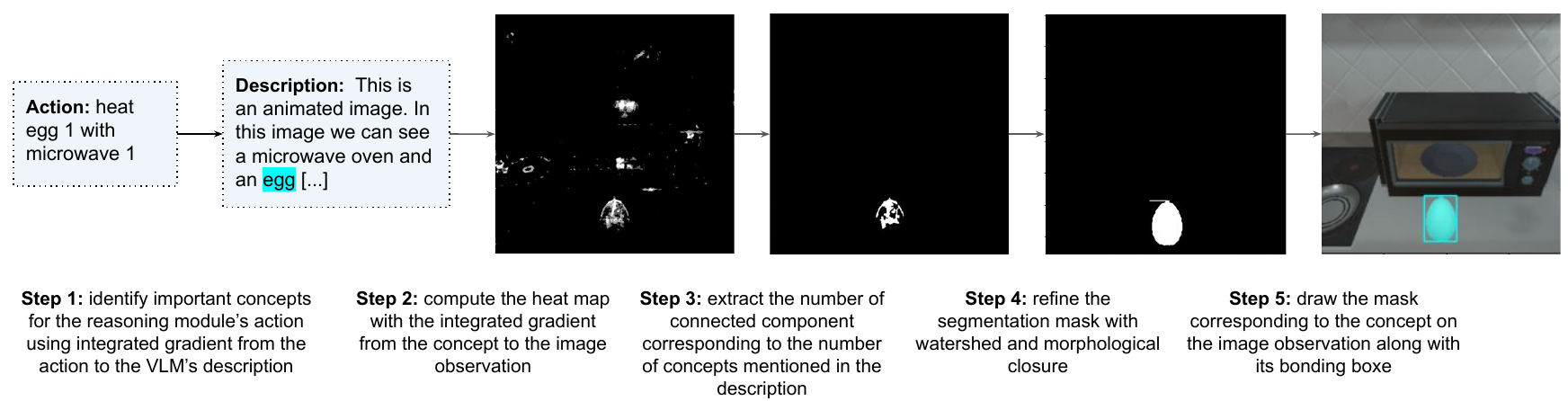}
    \caption{\textbf{Pipeline for Extracting Segmentation Masks of Key Instances for Decision-Making.}
Illustration of the step-by-step process used to identify important concepts influencing decision-making and localize their corresponding regions in the image.}
    \label{fig:mask_seg}
\end{figure*}

Beyond failure analysis, our explainability framework enables the identification of key concepts in \our intermediate textual descriptions and their corresponding locations in the image observations, as shown in Figure~\ref{fig:obs_sup}. As detailed in Figure~\ref{fig:mask_seg}, we first identify the important concepts in the VLM output description by computing the integrated gradients from the action selected by the LLM to the description generated by the VLM. For each important concept in the textual description (e.g. \textit{egg} in Figure~\ref{fig:mask_seg}), we then generate a heatmap using integrated gradients from the concept to the image. Next, we extract the top $n$ connected components from the heatmap, where $n$ corresponds to the number of concepts mentioned in the text (e.g. one for the egg). These components serve as priors for segmentation, and we apply the watershed algorithm to refine object boundaries, ultimately producing a final segmentation mask. The corresponding bounding box is then extracted from this mask.